\documentclass[a4paper,twoside]{article}

\usepackage{xparse}

\usepackage{type1cm}
\usepackage{color,soul}
\usepackage{epsfig}
\usepackage{subcaption}
\usepackage{calc}
\usepackage{amssymb}
\usepackage{amstext}
\usepackage{amsmath}
\usepackage{amsfonts}
\usepackage{amsthm}
\usepackage{multicol}
\usepackage{pslatex}
\usepackage{apalike}
\usepackage{markdown}
\usepackage{textcomp}
\usepackage{amsmath}
\usepackage{hyperref}
\usepackage{mathtools}
\usepackage{scalefnt}
\usepackage{cuted}
\usepackage{lipsum}
\usepackage[dvipsnames]{xcolor}
\usepackage[linesnumbered,ruled]{algorithm2e}
\usepackage[printonlyused, withpage, nohyperlinks]{acronym}
\usepackage{tabularx,array}
\usepackage{url}
\usepackage{bm}
\usepackage{booktabs}
\usepackage{tikz}
\usepackage{pgfplots}
\pgfplotsset{compat=newest}
\usepgfplotslibrary{groupplots}
\usepgfplotslibrary{dateplot}
\usetikzlibrary{arrows, positioning, shapes.geometric, automata, backgrounds, fit}
\usepackage{xspace}

\usepackage{SCITEPRESS}     % Please add other packages that you may need BEFORE the SCITEPRESS.sty package.

\newcommand\msmaller[2][0.76]{{\scalefont{#1}#2}}

\newcommand{\ifdeb}[2]{\ifdefined\debug{#2}\else{#1}\fi}

\newcommand{\comment}[1]{}
\newcommand{\mmark}[1]{{\begin{equation} abc \end{equation}}}

\mathchardef\mhyphen="2D

\newcommand{\nonnegintsset}{\mathbb{Z}_{\ge 0}}

\newcommand{\nphard}{{$\mathcal{NP}$-hard}\xspace}
\newcommand{\cit}[1]{\ifdeb{\cite{#1}}{Nobel}}
\newcommand{\jobsset}{\ensuremath{J}\xspace}
\newcommand{\jobssetrec}{\ensuremath{\jobsset}\xspace}
\newcommand{\pmax}{\ensuremath{j^{p\mhyphen max}}\xspace}
\newcommand{\dmin}{\ensuremath{j^{d\mhyphen min}}\xspace}
\newcommand{\authoretal}[1]{#1 \textit{ et al.}}
\newcommand{\authoretalcite}[2]{#1\textit{ et al.} \cit{#2}}

\newcommand{\n}{\ensuremath{n}\xspace}
\newcommand{\rdd}{\ensuremath{\mathit{rdd}}\xspace}
\newcommand{\tf}{\ensuremath{\mathit{tf}}\xspace}
\newcommand{\maxproc}{\ensuremath{p_{max}}\xspace}

\newcommand{\getss}{$\leftarrow$\xspace}

\DeclareMathOperator*{\argmin}{argmin}

\newcommand{\new}[1]{{#1}}
\newcommand{\job}{j}

\newcommand{\numjobs}{n}

\newcommand{\proctime}[1]{p_{#1}}

\newcommand{\duedate}[1]{d_{#1}}

\newcommand{\tardiness}[1]{T_{#1}}

\newcommand{\powerset}[1]{\ensuremath{\mathcal{P}({#1})}\xspace}

\newcommand{\position}{\ensuremath{k}\xspace}

\newcommand{\positionsset}{\ensuremath{K}}

\newcommand{\positionssetedd}{\ensuremath{\positionsset{}^{\edd}}\xspace}
\newcommand{\filpositionssetedd}{\ensuremath{\overline{\positionsset{}^{\edd}}}\xspace}
\newcommand{\positionssetspt}{\ensuremath{\positionsset{}^{\spt}}\xspace}
\newcommand{\filpositionssetspt}{\ensuremath{\overline{\positionsset{}^{\spt}}}\xspace}

\newcommand{\longestjobposition}{\ensuremath{\position^{*}}\xspace}

\ExplSyntaxOn
\DeclareExpandableDocumentCommand{\IfNoValueOrEmptyTF}{mmm}
 {
  \IfNoValueTF{#1}{#2}
   {
    \tl_if_empty:nTF {#1} {#2} {#3}
   }
 }
\ExplSyntaxOff

\DeclareDocumentCommand \pine { o o } {\IfNoValueOrEmptyTF{#1}{}{#2}}
\DeclareDocumentCommand \ifete { o o o } {\IfNoValueOrEmptyTF{#1}{#2}{#3}}

\DeclareDocumentCommand \sequence { o } {
    \IfNoValueTF {#1} {
      \pi
    }{
      \pi(#1)
    }
}

\newcommand{\before}[1]{\ensuremath{\mathit{before}}\xspace}

\newcommand{\after}[1]{\ensuremath{\mathit{after}}\xspace}

\newcommand{\precprobh}[2]{\ensuremath{P^{#1}\pine[#2][({#2})]\xspace}}

\newcommand{\follprobh}[2]{\ensuremath{F^{#1}\pine[#2][({#2})]\xspace}}

\newcommand{\precprob}[1]{\precprobh{}{#1}\xspace}

\newcommand{\follprob}[1]{\follprobh{}{#1}\xspace}

\newcommand{\precprobstar}[1]{\ensuremath{\precprob{#1}}}

\newcommand{\jed}[1]{\ensuremath{~\mathrm{#1}}}

\newcommand{\follprobstar}[1]{\ensuremath{\follprob{#1}}}

\newcommand{\precprobedd}[1]{\precprobh{\edd}{#1}}

\newcommand{\follprobedd}[1]{\follprobh{\edd}{#1}}

\newcommand{\precprobspt}[1]{\precprobh{\spt}{#1}}

\newcommand{\follprobspt}[1]{\follprobh{\spt}{#1}}

\newcommand{\obj}[1]{\ensuremath{Z\pine[#1][\left(#1\right)]}}
\newcommand{\predobj}[1]{\ensuremath{\widehat{\obj{}}\pine[#1][\left(#1\right)]}}

\newcommand{\inp}[1]{\ensuremath{\ifete[#1][\bm{X}][\bm{x}_{#1}]\xspace}}
\newcommand{\out}[1]{\ensuremath{y\pine[#1][_{#1}]}\xspace}

\acrodef{nlp}[NLP]{\emph{nature language processing}}
\newcommand{\NLP}{\ac{nlp}\xspace}
\acrodef{lstm}[LSTM]{\emph{Long Short-Term Memory \cit{Hochreiter1997}}}
\newcommand{\LSTM}{\ac{lstm}\xspace}
\acrodef{rnn}[RNN]{\emph{recurrent neural network}}
\newcommand{\RNN}{\ac{rnn}\xspace}
\acrodef{ml}[ML]{\emph{machine learning}}
\newcommand{\ML}{\ac{ml}\xspace}
\acrodef{dhs}[\textsc{horda}]{\emph{Heuristic Optimizer using Regression-based Decomposition Algorithm}}
\newcommand{\DHS}[1]{\ac{dhs}\pine[{#1}][\kern 0.06em+\kern 0.06em\textsc{\lowercase{{#1}}}]\xspace}
\acrodef{normSort}[\ensuremath{PSF}]{\emph{Preprocessing Sort Function}}

\acrodef{normConst}[\ensuremath{PC}]{\emph{Preprocessing Constant}}

\acrodef{normCritFunc}[\ensuremath{PCF}]{\emph{Preprocessing Criterion Function}}

\acrodef{normFeaFunc}[\ensuremath{PFF}]{\emph{Preprocessing Feature Function}}

\acrodef{ttbr}[\textsc{ttbr}]{Total Tardiness Branch-and-Reduce Algorithm}
\newcommand{\TTBR}[1]{\ac{ttbr}\pine[{#1}][\msmaller{\kern 0.09em{#1}s}]\xspace}
\acrodef{smttp}[SMTTP]{Single Machine Total Tardiness Problem}
\newcommand{\SMTTP}{\ac{smttp}\xspace}
\acrodef{tsp}[TSP]{Traveling Salesman Problem}
\newcommand{\TSP}{\ac{tsp}\xspace}
\acrodef{edd}[edd]{earliest due date}
\newcommand{\edd}{\ensuremath{edd}\xspace}
\acrodef{spt}[spt]{shortest processing time}
\newcommand{\spt}{\ensuremath{spt}\xspace}
\acrodef{or}[OR]{operations research}
\newcommand{\OR}{\ac{or}\xspace}
\newcommand{\NBR}{\textsc{nbr}\xspace}
\newcommand{\PSK}{\textsc{psk}\xspace}

\begin{document}

\title{Data-driven Algorithm for Scheduling with Total Tardiness}

\author{\authorname{
Michal Bou{\v s}ka\sup{1,2}\orcidAuthor{0000-0002-8034-2531},
Anton{\' i}n Nov{\' a}k\sup{1,2}\orcidAuthor{0000-0003-2203-4554},
P{\v r}emysl {\v S}{\r u}cha\sup{1}\orcidAuthor{0000-0003-4895-157X}, 
Istv{\' a}n M{\' o}dos\sup{1,2}\orcidAuthor{0000-0003-4692-1625},
and Zden{\v e}k Hanz{\' a}lek\sup{1}\orcidAuthor{0000-0002-8135-1296}}
\affiliation{\sup{1}Czech Institute of Informatics, Robotics and Cybernetics, Czech Technical University in Prague,\\ Jugosl\'{a}vsk\'{y}ch partyz\'{a}n\r{u} 1580/3, Prague,
Czech republic}
\affiliation{\sup{2}Czech Technical University in Prague, 
Faculty of Electrical Engineering, 
Department of Control Engineering, 
Karlovo~n\'{a}m\v{e}st\'{i} 13, Prague, Czech republic}
}


\abstract{
    In this paper, we investigate the use of deep learning for solving a classical \nphard{} single machine scheduling problem where the criterion is to minimize the total tardiness.
    Instead of designing an end-to-end machine learning model, we utilize well known decomposition of the problem and we enhance it with a data-driven approach.
    We have designed a regressor containing a deep neural network that learns and predicts the criterion of a given set of jobs.
    The network acts as a polynomial-time estimator of the criterion that is used in a single-pass scheduling algorithm based on Lawler\textquotesingle s decomposition theorem. 
    Essentially, the regressor guides the algorithm to select the best position for each job.
    The experimental results show that our data-driven approach can efficiently generalize information from the training phase to significantly larger instances (up to 350 jobs) where it achieves an optimality gap of about 0.5\%, which is four times less than the gap of the state-of-the-art \textsc{nbr} heuristic.
}

\onecolumn \maketitle \normalsize \setcounter{footnote}{0} \vfill

\section{\MakeUppercase{Introduction}}
% \idea{Data driven scheduling}
% \comment{
% \begin{markdown}
% * problem statement
% * motivation for the data-driven approach
%     * obchází lidskou práci učením z velkého množství dat
%     * velká časová náročnost na začátku, následně rychlá evaluace
% * related work
%     * OR pro SMTTP
%     * Data driven approach
%     * regresory/NN
% * contribution
%     * kombinace OR a ML
%     * LHS+NN poráží SOTA
%     * schopné řešit velké instance, oproti \TSP s 50V
%     * přepínání EDD, SPT
% * outline 
% \end{markdown}}
% Macro test:
% 0 argue: \pine \\
% 1 argue: \pine[a] \\
% 2 argue: \pine[a][acd] \\
% $\obj{}$ $Z$ \\
% $\obj{\jobsset}$ $Z(J)$ \\
% \predobj{\jobsset}
% \inp{} input\{\}\\
% \inp{1} input\{1\}\\
% \inp{11} input\{11\}\\

    The classical approaches for solving combinatorial problems have several undesirable properties.
    First, solving instances of an $\mathcal{NP}$-Hard problem to optimality consumes an unfruitful amount of computational time.
    Second, there is no well-established method how to utilize the solved instances for improving the algorithm or recycling the solutions for the unseen instances.
    Finally, the development of efficient heuristic rules requires a substantial time devoted to the research.
    To address these issues, we investigate the use of deep learning which is able to derive knowledge from the already solved instances of a classical scheduling \nphard{} \SMTTP and estimate the optimal value of an unseen \SMTTP instance.
    This is the first successful application of deep learning to the scheduling problem; we successfully integrated the deep neural network into a known decomposition algorithm and outperformed the state-of-the-art heuristics.
    With this, we are able to solve instances with hundreds of jobs, which is significantly more than, e.g., an end-to-end approach \cit{vinyals2015} that solves Traveling Salesman Problem with about 50 nodes.
    Our proposed approach outperforms the state-of-the-art heuristic for \SMTTP.
    \subsection{Problem Statement}
    \label{sec:problem}
        The combinatorial problem studied in this paper is denoted as $1||\sum \tardiness{\job}$ in Graham's notation of scheduling problems~\cit{graham1979}.
        Let $\jobsset = \{1,\dots,\numjobs\}$ be a set of jobs that has to be processed on a single machine.
        The machine can process at most one job at a time, the execution of the jobs cannot be interrupted, and all the jobs are available for processing at time zero.
        Each job $\job \in \jobsset$ has processing time $\proctime{\job} \in \nonnegintsset$ and due date $\duedate{\job} \in \nonnegintsset$.
        Let $\sequence: {\{1,\dots,\numjobs\}} \mapsto {\{1,\dots,\numjobs\}}$ be a bijective function representing a sequence of the jobs, i.e., $\sequence[\position] \in \jobsset$ is the job at position $\position$ in sequence $\sequence$.
        For a given sequence $\sequence$, tardiness of job $\sequence[\position]$ is defined as $\tardiness{\sequence[\position]} = \max\left(0, \left(\sum_{\position^{\prime} = 1}^{\position} \proctime{\sequence[\position^{\prime}]}\right) - \duedate{\sequence[\position]}\right)$.
        The goal of the scheduling problem is to find a sequence which minimizes the total tardiness, i.e., $\sum_{\job \in \jobsset} \tardiness{\job}$.
        The problem is proven to be \nphard{}~\cit{du1990}.
                
        In the rest of the paper, we use the following two definitions to describe the ordering of the jobs:
        \begin{enumerate}
            \item \ac{edd}: if $1 \le \job < \job^{\prime} \le \numjobs$ then either (i) $\duedate{\job} < \duedate{\job^{\prime}}$ or (ii) $\duedate{\job} = \duedate{\job^{\prime}} \, \wedge \,\proctime{\job} \le \proctime{\job^{\prime}}$,
            \item \ac{spt}: if $1 \le \job < \job^{\prime} \le \numjobs$ then either (i) $\proctime{\job} < \proctime{\job^{\prime}}$ or (ii) $\proctime{\job} = \proctime{\job^{\prime}} \, \wedge \,\duedate{\job} \le \duedate{\job^{\prime}}$.
        \end{enumerate}

    \subsection{Contribution and Outline}
        This paper addresses a single machine total tardiness scheduling problem using a machine learning technique.
        Unlike some existing works, for example, \cit{vinyals2015}, we do not purely count on machine learning, but we combine it with the known approaches from OR domain.
        The advantage of our approach is that it can extract specific knowledge from data, i.e., already solved instances, and use it to solve the new ones.
        The experimental results show two important observations.
        First, our algorithm outperforms the state-of-the-art heuristic \cit{holsenback1992}, and it also provides better results on some instances than the exact state-of-the-art approach \cit{Garraffa2018} with a time limit.
        Second, the proposed algorithm is capable of generalizing the acquired knowledge to solve instances that were not used in the training phase and also significantly differ from the training ones, e.g., in the number of jobs or the maximal processing time of jobs.

        The rest of the paper is structured as follows. In \autoref{sec:related-work}, we present a review of literature for \SMTTP and combination of \OR and \ML. 
        \autoref{sec:dec} describes our approach integrating a regressor into the decomposition and analyzes it's time complexity.
        We present results for standard benchmark instances for \SMTTP in \autoref{sec:experiments}.
        Finally, the conclusion is drawn in \autoref{sec:conclusion}.
        % Paper consists of the following section, review of literature for \SMTTP and combination of \OR and \ML, description of our approach, experimental results, and conclusion.
        
\section{\MakeUppercase{Related Work}}
    \label{sec:related-work}
    The first part of the literature overview is based on the extensive survey addressing \SMTTP published by \cit{koulamas2010} which we further extend with the description of the current state-of-the-art algorithms.
    The second part maps existing work in machine learning related to solving combinatorial problems.
    %There is a large volume of published studies investigating exact algorithms for the single machine total tardiness problem.

    % \todo{neches vytvorit subsubsection v ramci related work popisujici jednotlive casti? Ten skok z metaheuristik pro rozvrhovani na ML modely je ohromny a pritom je to vsechno v ramci jednoho textu v jedne sekci}
    \subsection{\SMTTP}
    
    In 1977 it was shown by Lawler \cit{lawler1977} that the weighted single machine total tardiness problem is $\mathcal{NP}$-Hard.
    However, it took more than a decade to prove that the unweighted variant of this problem is $\mathcal{NP}$-Hard as well \cit{du1990}.

    Lawler \cit{lawler1977} proposes a pseudo-polynomial (in the sum of processing times) algorithm for solving \SMTTP. 
    The algorithm is based on a decomposition of the problem into subproblems. 
    The decomposition selects the job with the maximum processing time and tries all the positions following its original position in the \edd order.
    For each position, two subproblems are generated; the first subproblem contains all the jobs preceding the job with the maximum processing time and the second subproblem contains all the jobs following the job with the maximum processing time.
    In addition, Lawler introduces rules for filtering the possible positions of the job with the maximum processing time. 
    This algorithm can solve instances with up to one hundred jobs.
    \authoretal{F. Della Croce} \cit{DellaCroce1998} proposed a \spt decomposition which selects the job with the minimal due date and tries all the positions preceding its original position in \spt order.
    Similarly as with the Lawler's decomposition, two subproblems are generated where the first subproblem contains all the jobs preceding the job with the minimal due date time and the second subproblem contains all the jobs following the job with the minimal due date.
    \authoretal{F. Della Croce} combined both \edd decomposition and \spt decomposition together, this presented algorithm is able to solve instances with up to 150 jobs.
    Finally, \authoretal{Szwarc} \cit{szwarc1999}  \new{integrate} the double decomposition from \cit{DellaCroce1998} and a split procedure from \cit{szwarc1996}.
    This algorithm was the state-of-the-art method for a long time with the ability to solve instance with up to 500 jobs.
    
    Recently, \authoretal{Garraffa} \cit{Garraffa2018} proposed \TTBR{}, which infers information about nodes of the search tree and merges nodes related to the same subproblem.
    This is the fastest known exact algorithm for \SMTTP to this date and is able to solve instances with up to 1300 jobs.

    Exact algorithms, such as the ones mentioned above, have very large computation times while the optimal solution is rarely needed in practice. 
    Hence, heuristic algorithms are often more practical.
    Existing heuristics algorithm can be categorized into the following three major groups.
    
    The first group of heuristics creates a job order and schedule the jobs according this order, i.e., list scheduling algorithms. There are various methods for creating a job order.
    The easiest one is to sort job by Earliest Due Date rule (\edd).
    A more efficient algorithm called \NBR{} was proposed in \cit{holsenback1992}. 
    \NBR{} is a constructive local search heuristic which starts with job set \jobsset sorted by \edd and constructs the schedule from the end by exchanging two jobs. 
    \authoretalcite{Panwalkar}{panwalkar1993} proposes constructive local search heuristic \PSK{}, which starts with job set \jobsset sorted by \spt and  constructs the schedule from the start by exchanging two jobs. 
    Russel and Holsenback \cit{russell1997b} compares \PSK{} and \NBR{} heuristic, and conducted that neither heuristic is inferior to another one. However, \NBR{} finds a better solutions in more cases. 
    The second group of heuristics is based on Lawler decomposition rule \cit{lawler1977}. 
    In this case, heuristic evaluates each child of the search tree node and the most promising child is expanded. This heuristic approach is evaluated in \cit{potts1991} with \edd heuristic as a guide for the search.
    The third group of heuristics are metaheuristics. 
    \cit{potts1991}, \cit{antony1996}, \cit{ben1996} present simulated annealing algorithm for \SMTTP.
    Genetic algorithms applied to \SMTTP are described in \cit{dimopoulos1999}, \cit{suer2012}, whereas \cit{bauer1999}, \cit{cheng2009} propose to use ant colony optimization for this scheduling problem.
    All the reported results in the previous studies are for instance sizes up to 100 jobs.
    However, these instances are solvable by the current state-of-the-art exact algorithm in a fraction of second.
    
    \subsection{Machine Learning Integration to Combinatorial Optimization Problems}
    The integration of \ML to combinatorial optimization problems has several difficulties.
    As first, \ML models are often designed with feature vectors having predefined fixed size.
    On the other hand, instances of scheduling problems are usually described by a variable number of features, e.g., variable number of jobs.
    This issue can be addressed by recurrent networks and, more recently, by encoder-decoder type of architectures.
    Vinyals \cit{vinyals2015} applied an architecture called Pointer Network that, given a set of graph nodes, outputs a solution as a permutation of these nodes.
    The authors applied the Pointer Network to \TSP, however, this approach for \TSP is still not competitive with the best classical solvers such as Concorde \cit{applegate2006concorde} that can find optimal solutions to instances with hundreds nodes in a fraction of second.
    Moreover, the output from the Pointer Network needs to be corrected by the beam-search procedure, which points out the weaknesses of this end-to-end approach.
    Pointer Network has achieved optimality gap around 1\% for instance with 20 nodes after performing beam-search.
    
    Second difficulty with training a \ML model is with acquisition of training data. 
    Obtaining one training instance usually requires solving a problem of the same complexity like the original problem itself.
    This issue can be addressed with reinforcement learning paradigm.
    \authoretal{Deudon} \cit{deudon2018learning} used encoder-decoder architecture trained with REINFORCE algorithm to solve 2D Euclidean \TSP with up to 100 nodes.
    It is shown that (i) repetitive sampling from the network is needed, (ii) applying well-known 2-opt heuristic on the results still improves the solution of the network, and (iii) both the quality and runtime are worse than classical exact solvers.
    Similar approach is described in~\cit{kool2018attention} which, if it is treated as a greedy heuristic, beats weak baseline solutions (from the operations research perspective) such as Nearest Neighbor or Christofides algorithm on small instances.
    To be competitive in terms of quality with more relevant baselines such as Lin-Kernighan heuristics, they perform multiple sampling from the model and output the best solution.
    Moreover, they do not directly compare their approach with state-of-the-art classical algorithms while admitting that off-the-shelf Integer Programming solver Gurobi solves optimally their largest instances within 1.5\jed{s}.
    
    \authoretal{Khalil} \cit{khalil2017learning} present an interesting approach for learning greedy algorithms over graph structures. The authors show that their S2V-DQN model can obtain competitive results on MAX-CUT and Minimum Vertex Cover problems. For \TSP, S2V-DQN performs about the same as 2-opt heuristics. Unfortunately, the authors do not compare runtimes with Concorde solver.
    
    \authoretal{Milan} \cit{milan2017data} presents a data-driven approximation of solvers for $\mathcal{NP}$-hard problems. They utilized a \LSTM network with a modified supervised setting. The reported results on the Quadratic Assignment Problem show that the network's solutions are worse than general purpose solver Gurobi while having the essentially identical runtime. 
    
    Integration of \ML with scheduling problems has received a little attention so far.
    Earlier attempts of integrating neural networks with job-shop scheduling are~\cit{zhou1991} and \cit{jain1998}.
    However, their computational results are inferior to the traditional algorithms, or they are not extensive enough to assess their quality.
    An alternative use of \ML in scheduling domain is focused on the criterion function of the optimization problems.
    For example, authors in~{\cit{vaclavik2016}} address a nurse rostering problem and improved the evaluation of the solutions' quality without calculating their exact criterion values.
    They propose a classifier, implemented as a neural network, able to determine whether a certain change in a solution leads to a better solution or not.
    This classifier is then used in a local search algorithm to filter out solutions having a low chance to improve the criterion function.
    Nevertheless, this approach is sensitive to changes in the problem size, i.e., the size of the schedule of nurses.
    If the size is changed, a new neural network must be trained.
    Another method, which does not directly predict a solution to the given instance, is proposed in~\cit{novak2015}.
    In this case, an online \ML technique is integrated into an exact algorithm where it acts as a heuristic.
    Specifically, the authors use regression for predicting the upper bound of a pricing problem in a Branch-and-Price algorithm.
    Correct prediction leads to faster computation of the pricing problem while incorrect prediction does not affect the optimality of the algorithm.
    This method is not sensitive to the change of the problem size; however, it is designed specifically for the Branch-and-Price approach and cannot be generalized to other approaches.
    \acresetall

\section{\MakeUppercase{Proposed Decomposition Heuristic Algorithm}}
\label{sec:dec}
In this section, we introduce \DHS{} for \SMTTP.
This heuristic effectively combines the well-know properties of \SMTTP and the data-driven approach.
Moreover, this paper proposes a methodology for designing data-driven heuristics for scheduling problems where good estimator of the optimization criterion can be obtained to guide the search.

This section is structured as follows.
First of all, we summarize decompositions used in the algorithm.
As the second, we describe \DHS{}.
Next we continue by discussing the architecture of the regressor, its integration into \SMTTP decompositions, and describe the training of the neural network.
Finally, we analyze the time complexity of \DHS{} algorithm.

\subsection{\SMTTP Decompositions}
    \label{sec:smttpdec}
    Firstly, we describe two different decomposition approaches for \SMTTP.
    The reason is that every state-of-the-art exact algorithm for \SMTTP is based on these two decompositions.
        
    First decomposition, introduced by Lawler \cit{lawler1977}, uses \edd (earliest due date) order in which it selects position for job \pmax{}, i.e., a job with the maximal processing time from job set \jobsset (in case of tie, \pmax is the job with the larger index in \edd order).
    %\todo{co takhle z toho udelat teorem? A timpaded muzes rict ze to je cele zalozene na tomhle teoremu (coz je pravda)}
    Lawler proves that there exists position $\position \in \{\pmax, \dots, \numjobs \}$ in the \edd order  such that at least one optimal solution exists where \pmax is preceded by all jobs $\{1,\dots,\position \} \setminus \{\pmax \}$ and followed by all jobs $\{\position + 1, \dots, \numjobs \}$.
    Let us denote set of positions $\{\pmax, \dots, \numjobs \}$ as $\positionssetedd$.
    %\todo{na zvazeni, nemela by mnozina pozic byt argumentovana J?}
    This property leads to the following exact decomposition algorithm.
    First, let $\precprobedd{} : \powerset{\jobsset} \times [1,\ldots,\n] \to \powerset{\jobsset}$ and $\follprobedd{}: \powerset{\jobsset} \times [1,\ldots,\n] \to \powerset{\jobsset}$ be functions  which for job set \jobsset and position \position return subproblem with jobs $\{1,\dots,\position \} \setminus \{ \pmax \}$ and $\{\position + 1, \dots, \numjobs \}$, respectively.
    Where \powerset{\jobsset{}} is powerset of \jobsset{}.
    % \todo{spravne by mela byt definice $P^{\textit{edd}}: \mathcal{P}(J) \times [1 .. n] \rightarrow \mathcal{P}(J)$, kde \( \mathcal{P}(J) \) je mnozina vsechn podmnozin \( J \).}
    Thus, for each eligible position $\position \in \{\pmax, \dots, \numjobs \}$, the problem is decomposed into two subproblems defined by \precprobedd{\jobsset, \position} and \follprobedd{\jobsset, \position} such that jobs \pmax is neither in \precprobedd{} nor in \follprobedd{}.
    Let $\obj{\jobsset}$ denote the optimal criterion value for job set $\jobsset$ computed as
    \begin{equation}
        \obj{\jobsset} = \min_{\position \in \positionssetedd} \obj{\jobsset, \position}\,,
    \end{equation}
    where
    % \begin{strip}
    \begin{equation}
        \label{eq:exact-criterion}
        \begin{split}
            \obj{\jobsset, \position} =
            \obj{\precprobedd{\jobssetrec,\position}} + \\
            \max\left(0, \proctime{\position} - \duedate{\position} + 
            % \smashoperator{
            \sum_{\job \in \precprobedd{\jobsset, \position}}
            % } 
            \proctime{\job}\right) + \\
            \obj{\follprobedd{\jobssetrec,\position}}\,.
        \end{split}
    \end{equation}
    % \end{strip}
    The optimal solution to the instance is found by recursively selecting the position \position with the minimal criterion \obj{}.
    
    The second decomposition\cit{DellaCroce1998} introduced by \authoretal{Della Croce} uses \spt order in which it selects position for job \dmin.
    We refer to this decomposition as \spt decomposition.
    Let us define \dmin job as a job with the minimal due date from job set \jobsset (in case of tie, \dmin is the job with the smaller index in the \spt order).
    Similarly as in the \new{\edd} decomposition proposed by Lawler, \authoretal{Della Croce} \cit{DellaCroce1998} prove that for job \dmin in \spt order there exists position $\position \in \{1, \dots, \dmin \}$  such that in at least one optimal solution \dmin is preceded by job set generated by function $\precprobspt{}: \powerset{\jobsset} \times [1,\ldots,\n] \to \powerset{\jobsset}$.
    \precprobspt{\jobsset, \position} returns job set with first \position jobs selected from $\{1, \dots, \dmin \}$ which are then sorted by \edd.
    Job \dmin is followed by job set $\follprobspt{}: \powerset{\jobsset} \times [1,\ldots,\n] \to \powerset{\jobsset}$ with all the others jobs.
    The set of positions $\position \in \{1, \dots, \dmin \}$ is denoted as $\positionssetspt$.
    One may use the \spt decomposition in the same recursive way as \edd decomposition to find the optimal solution.

    The efficiency of both decomposition approaches is significantly influenced by the branching factor.
    Here, the branching factor is equal to the number of eligible positions where job $\pmax \in \positionssetedd$ ($\dmin \in \positionssetspt$) can be placed.
    The number of eligible positions can be reduced by filtering rules described in \cit{lawler1977} and \cit{szwarc1999}.
    Let us denote that \filpositionssetedd, \filpositionssetspt are the sets \positionssetedd, \positionssetspt filtered by rules from \cit{szwarc1999} respectively.
    
\subsection{\DHS{}}
    Even though algorithms using decompositions proposed in \cit{lawler1977} and \cit{DellaCroce1998} are very efficient, their time complexity exponentially grows with the number of jobs.
    %The main challenge of any decomposition-based exact algorithm is an exponential growth of time complexity with the increasing depth of the search tree. 
    Our \DHS{} algorithm avoids this exponential growth by \new{pruning the search tree ruled by the polynomial-time estimation of \eqref{eq:exact-criterion} produced by a neural network.}
    The estimations of \( \obj{\jobsset, \position} \) and \( \obj{\jobsset}\) are denoted as \( \predobj{\jobsset, \position} \) and \( \predobj{\jobsset} \), respectively.

    \DHS{} algorithm is outlined in \autoref{code:dhs}.
    To increase the efficiency of the solution space search, our \DHS{} algorithm combines the power of both decompositions \cit{lawler1977} and \cit{DellaCroce1998} in the following way.
    The \DHS{} algorithm generates (lines \ref{code:dhs:genedd}~and~\ref{code:dhs:genspt}) two sets of eligible positions \filpositionssetedd and \filpositionssetspt by either \edd or \spt decomposition which are filtered by state-of-the-art rules \cit{szwarc1996}.
    Then, the set with the minimal cardinality is selected (lines \ref{code:dhs:selectstart}~-~\ref{code:dhs:selectend}) for the recursive expansion; we refer to the selected set as \positionsset.

    After obtaining positions set \positionsset{}, the algorithm greedily selects \longestjobposition position having the minimal estimation $\predobj{}$ (\autoref{code:dhs:argmin}). 
    Next, the algorithm recursively explores job sets \precprob{\jobssetrec,\longestjobposition} and \follprob{\jobssetrec,\longestjobposition}, and resulting partial sequences are stored as vectors \before{} and \after, respectively (lines \ref{code:dhs:recursionb}~and~\ref{code:dhs:recursiona}).
    Finally, the algorithm merges $\{\before{}, \longestjobposition, \after{}\}$ into one sequence, which is returned as the resulting schedule (\autoref{code:dhs:return}). 
    Note that job sets with less or equal than $5$ jobs are solved to optimality by an exact solver (\TTBR{}) instead of the decomposition.

    \begin{algorithm}[]
        \DontPrintSemicolon
        \KwData{\jobsset}
        \KwResult{\DHS{} ordered jobs}
        \SetKwFunction{FDHS}{\DHS{}}
        \SetKwProg{Fn}{Function}{:}{}
        \Fn{\FDHS{\jobssetrec}}{
     \If{$|\jobssetrec| \leq 1$} {
        \Return{toSequence(\jobssetrec)}
     }
     \tcc{Generate \edd and \spt positions with respect to the filtering rules}
     \positionssetedd \getss genEDDPos(\jobssetrec)\;\label{code:dhs:genedd}
     \positionssetspt \getss genSPTPos(\jobssetrec)\;\label{code:dhs:genspt}
     \If{$|\filpositionssetedd| \le |\filpositionssetspt|$\label{code:dhs:selectstart}}
     {
        \positionsset{} \getss \filpositionssetedd, \precprobstar{} \getss \precprobedd{}, \follprobstar{} \getss \follprobedd{}\;
     }\Else
     {
        \positionsset{} \getss \filpositionssetspt, \precprobstar{} \getss \precprobspt{}, \follprobstar{} \getss \follprobspt{}\;
     }\label{code:dhs:selectend}
        
     \tcc{Where \predobj{} is computed by regressor.}
     \longestjobposition \getss  $\argmin_{\position \in \positionsset}\,(\predobj{\precprobstar{\jobssetrec, \position}} + \max(0, \proctime{\position} - \duedate{\position} + \sum_{\job \in \precprobstar{\jobsset{}, \position}}\proctime{\job}) +  \predobj{\follprobstar{\jobssetrec, \position}})$\;  \label{code:dhs:argmin}

     \before{} \getss \FDHS{\precprobstar{\jobssetrec, \longestjobposition}}\; \label{code:dhs:recursionb}
     \after{} \getss \FDHS{\follprobstar{\jobssetrec, \longestjobposition}}\; \label{code:dhs:recursiona}
     \tcc{join sequences into one}
     order \getss (\before{}, \longestjobposition, \after{})\; \label{code:dhs:merge}
    \Return{order} \label{code:dhs:return}
    }
    \caption{Decomposition heuristic search (\DHS{})}
    \label{code:dhs}
\end{algorithm}

\subsection{Regressor}
    \label{sec:dec:regressor}
    The proposed \DHS{} algorithm utilizes the regressor estimation in the decomposition to guide the search by selecting position \longestjobposition that minimizes the estimated criterion \predobj{} (see \autoref{code:dhs:argmin}).
    The quality of the estimation significantly affects the quality of the found solutions.
    However, \DHS{} algorithm is not sensitive to absolute error of the estimation, instead, it's relative error is important.
    Therefore, the proposed regressor is based on neural networks that are known to be successful for problems sensitive to relative error, for example Google \cit{Silver2016} applied them to predict a policy in Monte Carlo Tree Search to solve game of Go.

    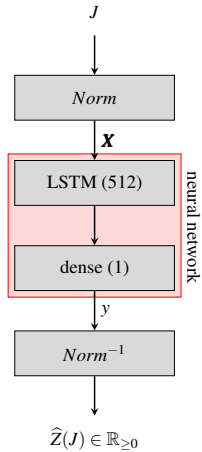
\begin{figure}
        \centering
        \scalebox{.7}{
        \begin{tikzpicture}[
            node distance = 0.75cm and 1.5cm,
            process/.style = {rectangle, draw, fill=gray!30,
                            minimum width=3cm, minimum height=0.85cm, align=center},
            data/.style = {rectangle, minimum height=0.85cm, align=center},
            arrow/.style = {thick,-stealth},
            bendarrow/.style = {thick,-stealth}
                                ]
            % nodes
            \node (n1)  [data] {\jobsset{}};
            \node (n2)  [process,below=of n1]   {$Norm$};
            \node (n3)  [process,below=of n2]   {LSTM (512)};
            \node (n4)  [process,below=of n3]   {dense (1)};
            \node (n5)  [process,below=of n4]   {$Norm^{-1}$};
            \node (n6)  [data,below=of n5]   {$\widehat{Z}(\jobsset) \in \mathbb{R}_{\geq 0}$};
            % connections
            \draw [arrow] (n1) -- (n2);
            \draw [arrow] (n2) -- (n3)node[midway,right] {\inp{}};
            %\draw [bendarrow] (n3) --  (n3); %[bend left,looseness=1.2]
            \draw [arrow] (n3) -- (n4)node[midway,right] {};
            \draw [arrow] (n4) -- (n5)node[midway,right] {\out{}};
            \draw [arrow] (n5) -- (n6);

            \begin{scope}[on background layer]
                \node[fit=(n3) (n4), rectangle, fill=red!30, draw=red, fill opacity=0.5, label={[label distance=-1.15cm,text depth=3ex,rotate=-90]right:neural network}] (qm) {};
            \end{scope}
        \end{tikzpicture}}
    \caption{Regressor architecture.}
    \label{tik:regressor}
    \end{figure}

    The architecture of our regressor using neural network is illustrated in \autoref{tik:regressor}.
    It has two main parts.
    The first one is the normalization of the input data, described in \autoref{sec:reg:preprocessing}.
    The second one is the neural network, explained in \autoref{sec:reg:nn}.

    \subsubsection{Input Data Preprocessing}
        \label{sec:reg:preprocessing}
        The speed of training and quality of the neural network is affected by the preprocessing of the input instances.
        There are two main reasons for the preprocessing denoted as $Norm$ in \autoref{tik:regressor}.
        Firstly, preprocessing of the input instance normalizes the instances, and thus reduces the variability of input data denoted \inp{}.
        For example, two neural network inputs differing only in job order are, in fact, the same.
        Secondly, numerical stability of the computation is improved by the preprocessing.
        In our regression architecture, the preprocessing has three main parts:
        \begin{enumerate}
            \item sorting of the input: we performed preliminary experiments with various sorting options such as \edd, \spt, reversed \edd and reversed \spt, among which \edd performed the best.
            \item normalization of the input: the processing times and due dates are divided by the sum of the processing times in the instance.
            \item appending additional features to the neural network: each job has one additional feature which is its position in \inp{} divided by the number of the jobs.    
        \end{enumerate}

        The best practice in the neural network training is to normalize value that is estimated by the neural network, denoted as \out{} in \autoref{tik:regressor}.
        In the training phase, the associated optimal criterion value of each instance is divided by the sum of the processing times.
        Alternatively, we evaluated one additional criterion normalization $\obj{} / \left(\n  \cdot \sum_{\job \in \jobsset}\proctime{\job}\right)$.
        However, it performed poorly.
        In the \DHS{} the estimation produced by the neural network has to be denormalized by the inverse transformation ($Norm^{-1}$ in \autoref{tik:regressor}) to obtain the actual estimation of the total tardiness.

    \subsubsection{Neural Network}
        \label{sec:reg:nn}
        The input data for our neural network have several similarities as the input data for \NLP problems.
        Firstly, as well as \NLP, our data can be arbitraly large, i.e., the size of job set \jobsset is unbounded; similarly, sentences in \NLP can be arbitrarily long.
        In other research fields, such as computer vision, this issue is mitigated by scaling the feature vectors to a fixed length.
        However, there is no simple and general way for scheduling problems how to aggregate multiple jobs into one without losing necessary information.
        Therefore, we use another technique of dealing with the varying length of the input which are \RNN{} \cit{Sundermeyer2012}.

        Our neural network for the criterion estimation of \jobsset{} consists of two parts (the red box in \autoref{tik:regressor}).
        The first layer is \LSTM, which receives job set \jobsset{} as the input.
        The input $\inp{}$ is a sequence of features $\inp{\job}$ for every job \( \job \in \jobsset\).
        Each feature vector $\inp{\job}$ consists of $\proctime{\job}$ and $\duedate{\job}$ with additional features described below.
        The output of the last LSTM step is passed into a dense layer which produces estimation \out{} of the criterion for \inp{}.

\subsection{Time Complexity of \DHS{}}
    \label{sec:dec:time-complexity}
    In this section, we present the worst-case runtime of \DHS{}.    
    The most time consuming part of \DHS{} is the estimation of \predobj{\jobsset} by the regressor.
    The  \LSTM layer produces \predobj{\jobsset} in $O(\n)$ time and \DHS{} algorithm evaluates the regressor $2 \cdot \n$ times to select position \longestjobposition from \positionsset{}.
    Thus, the evaluation of all the estimations for \positionsset{} takes $O(\n^2)$.
    In the worst-case, when decomposition repetitively removes one job, \DHS{} algorithm makes $O(\n)$ selections of position \longestjobposition.
    Therefore, the worst-case time complexity of \DHS{} algorithm is $O(n^3)$.
    However, we note that the constants present in the asymptotic complexit are fairly low. 
    Hence, it is efficient in practice, as well.

\section{\MakeUppercase{Experimental Results}}
\label{sec:experiments}
\begin{table}[b!]
    \centering
    \caption{Mean \TTBR{}\cit{Garraffa2018} runtimes in seconds with respect to instance parameters for $\n \in \{5,\ldots,500\}$ and $\maxproc=100$. For parameters relative range of due dates (\rdd), and the average tardiness factor (\tf).}
    \label{tab:tkind-time-rdd-tf}
    \begin{tabular}{l|l l l l l}
    \toprule
    $\rdd / \tf$ & 0.2 & 0.4 & 0.6 & 0.8 & 1.0 \\
    \midrule
    0.2 & 0.07 & 2.16 & \textbf{5.16} & 1.64 & 0.04 \\ 
    0.4 & 0.04 & 0.36 & 1.64 & 0.05 & 0.04 \\ 
    0.6 & 0.04 & 0.06 & 0.47 & 0.04 & 0.04 \\ 
    0.8 & 0.04 & 0.04 & 0.07 & 0.04 & 0.04 \\ 
    1.0   & 0.04 & 0.04 & 0.04 & 0.04 & 0.04 \\ 
    \bottomrule
    \end{tabular}
\end{table}
\begingroup
\setlength{\tabcolsep}{5pt}
\begin{table*}[]
\vspace{5pt}
    \centering
    \caption{Optimality gap of \DHS{}, \TTBR{}\cit{Garraffa2018} and \NBR{}\cit{holsenback1992} on instances with $\maxproc = 100$.}
    \label{tab:nbr-dhs}
    \begin{tabular*}{\textwidth}{l | @{\extracolsep{\fill}} r | c c | c c | c r} 
    \toprule
    $n$ & \multicolumn{1}{c}{\textbf{\TTBR{}}} & \multicolumn{2}{c}{\textbf{\NBR{}}} & \multicolumn{2}{c}{\textbf{\DHS{\NBR{}}}} & \multicolumn{2}{c}{\textbf{\DHS{NN}}} \\
    $\pm25$ & \multicolumn{1}{c}{time [s]} & \multicolumn{1}{c}{gap [\%]} & \multicolumn{1}{c}{time [s]} &\multicolumn{1}{c}{gap [\%]} & \multicolumn{1}{c}{time [s]} & \multicolumn{1}{c}{gap [\%]} & \multicolumn{1}{c}{time [s]} \\
    \midrule
        225 & $1.05\pm2.90$ & $1.98\pm0.58$ & $0.06\pm0.01$ & $1.17\pm0.47$ & $1.19\pm0.42$ & $\textbf{0.58}\pm0.30$ & $5.03\pm8.16$\\
        275 & $2.45\pm4.19$ & $2.12\pm0.54$ & $0.09\pm0.02$ & $1.31\pm0.44$ & $1.91\pm0.62$ & $\textbf{0.57}\pm0.28$ & $6.89\pm9.62$\\
        325 & $4.72\pm4.09$ & $2.20\pm0.50$ & $0.12\pm0.02$ & $1.39\pm0.43$ & $2.87\pm0.90$ & $\textbf{0.57}\pm0.37$ & $9.25\pm11.29$\\
        375 & $8.42\pm4.75$ & $2.27\pm0.49$ & $0.17\pm0.03$ & $1.46\pm0.44$ & $4.15\pm1.31$ & $\textbf{1.23}\pm0.63$ & $14.61\pm13.52$\\
        425 & $14.42\pm8.06$ & $2.34\pm0.46$ & $0.21\pm0.04$ & $\textbf{1.55}\pm0.41$ & $5.52\pm1.71$ & $1.71\pm0.65$ & $20.60\pm17.00$\\
    \bottomrule
    \end{tabular*}
\end{table*}
\endgroup
\begin{table*}[]
    \centering
    \caption{Optimality gap of heuristics on instances with $\maxproc = 5000.$}
    \label{tab:tkindt-dhs-5000}
    \begin{tabular*}{\textwidth}{l | @{\extracolsep{\fill}} r | c | c | c | c r}
    \toprule
    $n$ & \multicolumn{1}{c}{\textbf{\TTBR{}}} & \multicolumn{1}{c}{\textbf{\TTBR{10}}} & \multicolumn{1}{c}{\textbf{\NBR{}}} & \multicolumn{1}{c}{\textbf{\DHS{\NBR{}}}} & \multicolumn{2}{c}{\textbf{\DHS{NN}}} \\
    $\pm25$ & \multicolumn{1}{c}{time [s]} & \multicolumn{1}{c}{gap [\%]} & \multicolumn{1}{c}{gap [\%]} &\multicolumn{1}{c}{gap [\%]} & \multicolumn{1}{c}{gap [\%]} & \multicolumn{1}{c}{time [s]} \\
    \midrule
    225 & $10.66\pm9.20$ & $\textbf{0.17}\pm0.31$ & $1.91\pm0.60$ & $1.10\pm0.48$ & $0.58\pm0.27$ & $3.58\pm0.81$\\
    275 & $40.36\pm32.24$ & $0.77\pm0.69$ & $2.00\pm0.54$ & $1.20\pm0.45$ & $\textbf{0.55}\pm0.27$ & $4.89\pm1.02$\\
    325 & $92.30\pm56.39$ & $1.28\pm0.86$ & $2.27\pm0.53$ & $1.36\pm0.47$ & $\textbf{0.53}\pm0.33$ & $6.61\pm1.50$\\
    375 & $212.69\pm122.14$ & $1.87\pm0.87$ & $2.39\pm0.47$ & $1.50\pm0.48$ & $\textbf{1.09}\pm0.60$ & $10.32\pm2.18$\\
    425 & $488.76\pm265.88$ & $2.64\pm0.87$ & $2.32\pm0.44$ & $\textbf{1.52}\pm0.41$ & $1.73\pm0.64$ & $14.96\pm2.00$\\       
    \bottomrule
    \end{tabular*}
\end{table*}

    In this section, we present the experimental results. Firstly, we describe the training of the neural network, also with the acquisition of a training dataset. Secondly, we describe the generation of the benchmark instances.
    Then we compare our \DHS{} heuristic with the state-of-the-art heuristic \NBR{} \cit{holsenback1992} and exact algorithm \TTBR{} ~ \cit{Garraffa2018}.
    Finally, we discuss the advantages of our proposed heuristic.

    Experiments were run on a single-core of the Xeon(R) Gold 6140 processor with a memory limit set to 8GB of RAM.
    \DHS{} and \NBR{} algorithms were implemented in Python, and the neural network is trained in Tensor Flow 1.14 on Nvidia GTX 1080 Ti.
    Source codes of \TTBR{} algorithm were provided by authors of \cit{Garraffa2018} and it is implemented in C.

    \subsection{Neural Network Training}
    \label{sec:res:nn-training}
    We trained the neural network with Adam optimizer, with learning rate set to 0.0001, early stop with patience equals to 5.
    Size of the \LSTM layer is set to 512.
    For the neural network training, we generated instances by scheme introduced by Potts and Wassenhove~\cite{Potts1982}.
    The scheme uses two parameters;  relative range of due dates (\rdd), and the average tardiness factor (\tf).
    The values of \rdd, \tf typically used in the literature are $\rdd, \tf \in \{0.2, 0.4, 0.6, 0.8, 1\}$.
    For each such \rdd, \tf and $\numjobs \in \{5,\ldots,250\}$, we generated 5000 instances.
    Therefore, the whole training dataset consists of 30625000 instances in total.
    Since we use a supervised learning to train the neural network, we need optimal criterion values that acts as labels.

    It is easy to see that the dataset is enormous, and it is necessary to solve millions of \SMTTP instances.
    However, this is not an issue since a substantial amount of the instances can be solved within a fraction of a second.
    Moreover, the dataset can be cheaply generated in the cloud, e.g., on the Amazon EC2 cloud, the cost of generating the dataset is around $800\$$ and takes only ten days, which is significantly cheaper compared to the cost of a human expert developing a heuristic algorithm.

    Furthermore, it is important to stress that our neural network is able to generalize to larger instance than used in the training.
    Therefore, it is possible to train the neural network on smaller instances and solve larger ones both in terms of the number of jobs and their parameters.
    % \todo{Asi bych přidal nějaké krátké hodnocení trénování NN?}

    \subsection{Benchmark Instances}
    \begin{figure*}[h!]
        \centering
        \input{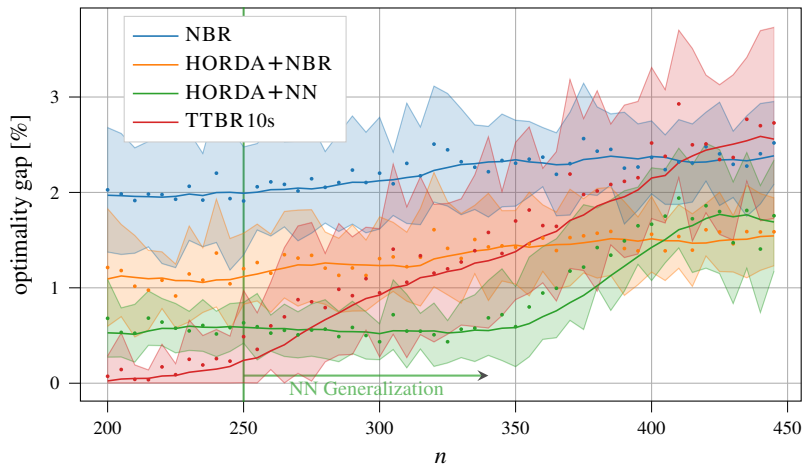}
        \caption{Optimality gap on instances with $\maxproc = 5000.$}
        \label{fig:tkindt-dhs-5000}
    \end{figure*}
    Benchmark instances used in this paper were generated in the manner suggested by Potts and Van Wassenhove in \cit{potts1991} and used in \autoref{sec:res:nn-training}.
    Potts and Van Wassenhove generate processing times of jobs uniformly on the interval from $1$ to $100$.
    We define maximal processing time \maxproc and generate processing time of jobs in instance uniformly on the interval from $1$ to \maxproc.
    For $\maxproc = 100$ and $\n \in \{5,\ldots,500\}$, we generated 25 sets of benchmarks differing in \rdd and \tf.
    Then those instances were solved by \TTBR{} algorithm. 
    \autoref{tab:tkind-time-rdd-tf} shows average runtimes in seconds over $(\rdd, \tf)$ $\in \{0.2, 0.4, 0.6, 0.8, 1\}^2$.
    The results imply, that the hardest instances occur for $\rdd = 0.2$ and $\tf = 0.6$ \new{(highlighted in \autoref{tab:tkind-time-rdd-tf} in bold)}, therefore our experiments concentrate on them.
    Nevertheless, it is important to stress that the neural network is trained on the whole range of values $(\rdd, \tf)$.
    First, we do not want the algorithm to be limited to a specific class of instances.
    Second, since our algorithm uses the decompositions, as described in Section~\ref{sec:smttpdec}, there is no guarantee that the subproblems have the same $(\rdd, \tf)$ parameterization as the input instance.
    In fact, during the run of \DHS{}, the values of $(\rdd, \tf)$ in newly emerged subproblems shift from the original ones.

    \subsection{Comparison with Existing Approaches}
    In the first experiment, summarized in \autoref{tab:nbr-dhs}, we concentrate on the comparison with \NBR{} heuristic.
    The benchmark instances used in this experiment were generated with $\maxproc = 100$. Each row in the table represents a set of 200 instances of size from range $[n-25, n+25)$. The optimal solution was obtained by \TTBR{} algorithm.
    The table compares \NBR{} heuristic with \DHS{} algorithm where the regressor is substituted by \NBR{} heuristic (denoted \DHS{\NBR{}}), and \DHS{} heuristic with the neural network regressor (denoted \DHS{NN}).
    These three approaches are compared in terms of the average CPU time, and the average quality of solutions, measured by the optimality gap in percent.
    All values are reported together with their standard deviation.

    Results are shown from $\n = 200$.
    For smaller \n than $200$, \TTBR{} is able to find the optimal solution under a second, and because of this, the results of heuristics are not relevant.
    % \todo{jak je to vlastne s vysledkama pro mensi n? treba 30? Me prijde ze \NBR{} se zhorsuje s velikosti dat, tak pro male by to mohlo byt dobre.} 
    The bold values in the table indicate the best result over all the heuristic approaches for the particular set of instances.
    The results show that \DHS{NN} has the best performance in terms of the average optimality gap.
    In the case of the last data set, the second heuristic \DHS{\NBR{}} is slightly better.
    The reason is that the neural network was trained only on instances with $n \leq 250$.
    Therefore, one can see that our neural network, used in the regressor, is able to generalize the gained knowledge to instances with $n \leq 400$.
    On instances with $n \leq 325$, the average optimality gap of \DHS{NN} is about $0.5\%$, which outperforms all other methods. At the same time, we have to admit that the heuristic is slower than \TTBR{} algorithm.
    Nevertheless, this is true only on instances generated with $\maxproc = 100$.
    On larger maximum processing time, the CPU time of \TTBR{} is significantly larger as will be seen in the next experiment.

    In literature, benchmark instances for \SMTTP are usually generated with $\maxproc = 100$, as it was used in the previous experiment. 
    %However, Lawler shows that \SMTTP is pseudopolynomial with respect to maximal processing time \todo{problem neni pseudopolynomialni, algoritmus je pseudopolynomialni? Nechteli ste tady rict, ze TTBR je pseruodpoly kvuli tomu, ze uvnitr nejak resi to pseudopolynomialni dynamicke programovani?}. 
    Since  \SMTTP is applicable in production and grid computing and \maxproc can be much longer in these fields, we introduce the following experiments with maximal processing time \maxproc equal to 5000.
    \autoref{tab:tkindt-dhs-5000} compares our \DHS{NN} and \DHS{\NBR{}} heuristics with \NBR{}, \TTBR{} and \TTBR{} with runtime limited to 10\jed{s} denoted as \TTBR{10}.
    For \TTBR{10}, a 10\jed{s} limit is selected with respect to the \DHS{NN} algorithm runtime, since the runtime of \DHS{NN} on instances with up to $\n=350$ is under 10\jed{s}.
    Please note that the identical regressor as in Table~\ref{tab:nbr-dhs} was used, i.e., the regressor was trained only on instances with $\maxproc=100$.
    Hence, it demonstrates neural network's ability of generalization outside the training processing time range.

    One can observe from \autoref{tab:nbr-dhs} and \autoref{tab:tkindt-dhs-5000} that the CPU time of \DHS{NN} is almost the same for both types of instances.
    However, this is not true for \TTBR{} where the CPU time is almost 30 times higher for $\n = 425$.
    Also, the CPU time of \TTBR{} is more than 30 times higher for $n = 425$ and $\maxproc = 5000$ compared to \DHS{NN}.
    If the runtime of \TTBR{} is limited to 10\jed{s}, then \DHS{NN} outperforms \TTBR{10} on larger instances.
    Moreover, the optimality gap of \DHS{NN} is practically the same as in the previous experiment with $\maxproc = 100$.

    The same experiment is shown in the form of a graph in \autoref{fig:tkindt-dhs-5000}. It compares the optimality gap of \NBR{}, \TTBR{} with a time limit, \DHS{\NBR{}}, and \DHS{NN}.
    The bold lines in the graph represent the moving average (last 5 samples) of optimality gap of each method, and the colored areas represent their standard deviation.
    \DHS{NN} outperforms \DHS{\NBR{}} about two times up to instances of size $\n=360$. 
    For instances with $n \geq 405$, \DHS{\NBR{}}, is slightly better. 
    In addition, \DHS{NN} also outperforms \TTBR{10} from $\n=265$.
    Furthermore, \DHS{NN} holds the average optimality gap around 0.5\% for instances with up to $350$ jobs.
    The same can be observed on instances with $\maxproc = 100$ (see \autoref{tab:nbr-dhs}).
    Finally, the runtime of \TTBR{} grows exponentially with the growing size of the instance, in contrast to polynomial runtime of \DHS{NN}.

    Concerning the heuristic using the neural network (\DHS{NN}), it is important to stress that for instances with $n > 250$ the network has to generalize the acquired knowledge since it was trained only on instances with $n \leq 250$.
    This fact is indicated in \autoref{fig:tkindt-dhs-5000} by a green vertical line.
    It can be seen that \DHS{NN} is able to generalize results to instances having 100 more jobs than instances encountered in the training phase with 50 times larger maximal processing time (instances for the training phase were generated with $\maxproc = 100$).

\section{\MakeUppercase{Conclusion}}
    \label{sec:conclusion}
    To the best of our knowledge, this is the first paper addressing a scheduling problem using deep learning.
    Unlike the solution used in~\cite{vinyals2015}, which tackled the Traveling Salesman Problem, we combined a state-of-the-art operations research method with a DNN.
    The experimental results show that our approach provides near-optimal solutions very quickly and is also able to generalize the acquired knowledge to larger instances without significantly affecting the quality of the solutions.
    Our approach outperforms state-of-the-art heuristic \NBR{}.
    Our approach is shown to be competitive and in some cases, superior to the previous state-of-the-art algorithms.
    Hence, we believe that the proposed methodology opens new possibilities for the design of efficient heuristics algorithms.
    % One may argue that the data-driven approaches like this one or~\cite{vinyals2015} cannot pay off since the time needed to train the DNNs is not negligible and should be included into the time required to solve a problem instance.
    % Although this is a valid line of reasoning, in many situations it is possible to invest computational effort into training accurate models offline and use them later to find solutions when they are quickly needed.
    
\section*{\uppercase{Acknowledgements}}
The authors want to thank Vincent T'Kindt from Université de Tours for providing the source code of \TTBR{} algorithm.
    This work was supported by the European Regional Development Fund under the project AI\&Reasoning (reg. no. CZ.02.1.01/0.0/0.0/15\_003/0000466).
    
This work was supported by the Grant Agency of the Czech Technical University in Prague, grant No. SGS19/175/OHK3/3T/13.

\bibliographystyle{apalike}
{\small
\bibliography{example}}

% \section*{\uppercase{Appendix}}

% \noindent If any, the appendix should appear directly after the
% references without numbering, and not on a new page. To do so please use the following command:
% \textit{$\backslash$section*\{APPENDIX\}}

\end{document}